\documentclass[runningheads]{llncs}

\usepackage[T1]{fontenc}
\usepackage{graphicx}
\usepackage{xcolor}
\usepackage{amsfonts} 
\usepackage{amsmath}
\DeclareMathOperator*{\argmin}{arg\,min} 
\usepackage{soul}
\usepackage{multirow}

\usepackage{hyperref}
\usepackage{color}

\begin{document}

\title{ProtoASNet: Dynamic Prototypes for Inherently Interpretable and Uncertainty-Aware Aortic Stenosis Classification in Echocardiography}
\titlerunning{ProtoASNet: Interpretable and Uncertainty-Aware AS Classification}

\author{
Hooman Vaseli\inst{1} \and  
Ang Nan Gu\inst{1} \and 
S. Neda Ahmadi Amiri  \inst{1} \and
Michael Y. Tsang \inst{2} \and 
Andrea Fung \inst{1} \and 
Nima Kondori \inst{1} \and 
Armin Saadat\inst{1}\and 
Purang Abolmaesumi \inst{1} \and 
Teresa S. M. Tsang \inst{2} 
}

\authorrunning{H. Vaseli et al.}

\institute{Department of Electrical and Computer Engineering, The University of British Columbia, Vancouver, BC, Canada \\ \email{\{hoomanv,guangnan,purang\}@ece.ubc.ca}
\and Vancouver General Hospital, Vancouver, BC, Canada 
\footnote{T. S.M. Tsang and P. Abolmaesumi are joint senior authors.
\\ H. Vaseli, A. Gu, and N. Ahmadi are joint first authors.} 
}

\maketitle

\begin{abstract}
Aortic stenosis (AS) is a common heart valve disease that requires accurate and timely diagnosis for appropriate treatment. 
Most current automatic AS severity detection methods rely on black-box models with a low level of trustworthiness, which hinders clinical adoption. To address this issue, we propose ProtoASNet, a prototypical network that directly detects AS from B-mode echocardiography videos, while making interpretable predictions based on the similarity between the input and learned spatio-temporal prototypes. This approach provides supporting evidence that is clinically relevant, as the prototypes typically highlight markers such as calcification and restricted movement of aortic valve leaflets. Moreover, ProtoASNet utilizes abstention loss to estimate aleatoric uncertainty by defining a set of prototypes that capture ambiguity and insufficient information in the observed data. This provides a reliable system that can detect and explain when it may fail. We evaluate ProtoASNet on a private dataset and the publicly available TMED-2 dataset, where it outperforms existing state-of-the-art methods with an accuracy of $80.0\%$ and $79.7\%$, respectively. Furthermore, ProtoASNet provides interpretability and an uncertainty measure for each prediction, which can improve transparency and facilitate the interactive usage of deep networks to aid clinical decision-making. Our source code is available at: \url{https://github.com/hooman007/ProtoASNet}.

\keywords{Aleatoric Uncertainty  \and Aortic Stenosis \and Echocardiography \and Explainable AI \and Prototypical Networks}
\end{abstract}

\section{Introduction}
Aortic stenosis (AS) is a common heart valve disease characterized by the calcification of the aortic valve (AV) and the restriction of its movement.
It affects 5\% of individuals aged 65 or older~\cite{ancona2020epidemiology} and can progress rapidly from mild or moderate to severe, reducing life expectancy to 2 to 3 years~\cite{thoenes2018patient}.
Echocardiography (echo) is the primary diagnostic modality for AS. This technique measures Doppler-derived clinical markers~\cite{acc_guidelines20212020} and captures valve motion from the parasternal long (PLAX) and short axis (PSAX) cross-section views. 
However, obtaining and interpreting Doppler measurements requires specialized training and is subject to significant inter-observer variability~\cite{minners2008inconsistencies,minners2010inconsistent}.

To alleviate this issue, deep neural network (DNN) models have been proposed for automatic assessment of AS directly from two-dimensional B-mode echo, a modality more commonly used in point-of-care settings.
Huang et al.~\cite{huangSemisupervisedEchocardiogramBenchmark2021,huangTMED2Dataset2022} proposed a multitask model to classify the severity of AS using echo images.
Ginsberg et al.~\cite{AS_Tom} proposed an ordinal regression-based method that predicts the severity of AS and provides an estimate of aleatoric uncertainty due to uncertainty in training labels.
However, these works utilized black-box DNNs, which could not provide an explanation of their prediction process.

Explainable AI (XAI) methods can provide explanations 
of a DNN's decision making process and can generally be categorized into two classes.
Post-hoc XAI methods explain the decisions of trained black-box DNNs. 
For example, gradient-based saliency maps~\cite{GradCam,journals/corr/SimonyanVZ13} show where a model pays attention to, but these methods do not necessarily explain why one class is chosen over another~\cite{rudin2019stop}, and at times result in misleading explanations~\cite{adebayo2018sanity}.
Ante-hoc XAI methods are explicitly designed to be explainable. 
For instance, prototype-based models~\cite{chen2019ProtoPNet,hesse2022insightr,XAI_RegionGrouping,XprotoNet,Trinh_2021_WACV,wang2021TesNet}, which the contributions of our paper fall under, analyze a given input based on its similarity to learned discriminative features (or ``prototypes'') for each class. 
Both the learned prototypes and salient image patches of the input can be visualized for users to validate the model's decision making. 

There are two limitations to applying current prototype-based methods to the task of classifying AS severity from echo cine series.
First, prototypes should be spatio-temporal instead of only spatial, since AS assessment requires attention to small anatomical regions in echo (such as the AV) at a particular phase of the heart rhythm (mid-systole).
Second, user variability in cardiac view acquisition and poor image quality can complicate AV visualization in standard PLAX and PSAX views. 
The insufficient information in such cases can lead to more plausible diagnoses than one. 
Therefore, a robust solution should avoid direct prediction and notify the user.
These issues have been largely unaddressed in previous work.

\begin{figure}[t]
\centering
\includegraphics[width=0.99\textwidth]{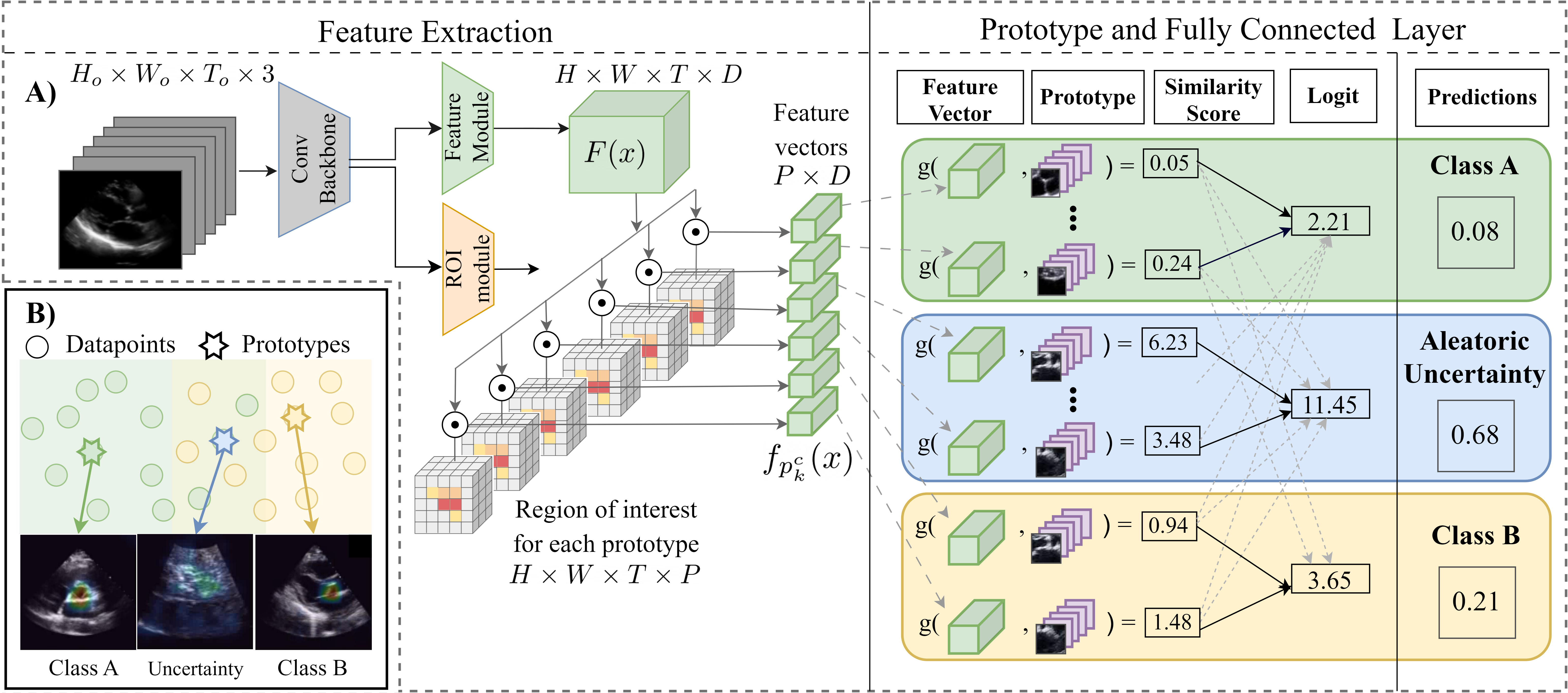}
\caption{\textbf{(A)} An overview of our proposed ProtoASNet architecture. ProtoASNet extracts spatio-temporal feature vectors $f_{p_k^c}(x)$ from the video, which are compared with learned prototypes. Similarity values between features and prototypes are aggregated to produce a score for class membership and aleatoric uncertainty.
\textbf{(B)} Prototypes representing aleatoric uncertainty (blue) can capture regions of the data distribution with inherent ambiguity (intersection between green and yellow regions). In practice, this region consists of videos with poor visual quality.}
\label{fig:network}
\end{figure}

We propose ProtoASNet (Fig.~\ref{fig:network}), a prototype-based model for classifying AS severity from echo cine series. 
ProtoASNet discovers dynamic prototypes that describe shape- and movement-based phenomena relevant to AS severity, outperforming existing models that only utilize image-based prototypes. 
Additionally, our model can detect ambiguous decision-making scenarios based on similarity with less informative samples in the training set. This similarity is expressed as a measure of aleatoric uncertainty.
To the best of our knowledge, the only prior work for dynamic prototypes published to-date is \cite{gulshad2023hierarchical}. 
ProtoASNet is the first work to use dynamic prototypes in medical imaging and the first to incorporate aleatoric uncertainty estimation with prototype-based networks.

\section{Methods}

\subsection{Background: Prototype-Based Models}
Prototype-based models explicitly make their decisions using similarities to cases in the training set. 
These models generally consist of three key components structured as $h(g(f(x)))$.
Firstly, $f(.)$ is a feature encoder such as a ConvNet that maps images 
$x \in \mathbb{R} ^{H_o \times W_o \times 3}$ to $f(x) \in \mathbb{R} ^{H \times W \times D}$,
where $H$, $W$, and $D$ correspond to the height, width, and feature depth of the ConvNet's intermediate layer, respectively.
Secondly, $g(.) \in \mathbb{R} ^{H \times W \times D} \rightarrow \mathbb{R} ^{P}$ is a prototype pooling function that computes the similarity of encoded features $f(x)$ to $P$ prototype vectors. There are $K$ learnable prototypes defined for each of $C$ classes, denoted as $p_k^c$. 
Finally, $h(.) \in \mathbb{R} ^{P} \rightarrow \mathbb{R} ^{C}$ is a fully-connected layer that learns to weigh the input-prototype similarities against each other to produce a prediction score for each class.
To ensure that the prototypes $p_k^c$ reflect those of true examples in the training distribution, they are projected (``pushed'') towards the embeddings of the closest training examples of class $c$.
\begin{equation}
\label{eqn:proto_push}
     p_k^c \leftarrow \argmin_{z \in \mathcal{Z}_c}  \lVert z - p_k^c \rVert_2 , \text{where} \  \mathcal{Z}_c = \{z: z \in f_{p_k^c}(x_i) \ s.t. \  y_i \in c \}
\end{equation}
Such models are inherently interpretable since they are enforced to first search for similar cases in the training set and then to compute how these similarities contribute to the classification. As a result, they offer a powerful approach for identifying and classifying similar patterns in data.
\subsection{ProtoASNet}
\subsubsection{Feature Extraction.} The overall structure of ProtoASNet is shown in Fig.~\ref{fig:network}.
The feature extraction layer consists of a convolutional backbone, in our case the first three blocks of a pre-trained R(2+1)D-18~\cite{tran2018closer} model,
followed by two branches of feature and region of interest (ROI) modules made up of two and three convolutional layers respectively.
In both modules, the convolutional layers have ReLU activation function, except the last layers which have linear activations.
Given an input video $x \in \mathbb{R} ^{H_o \times W_o \times T_o \times 3}$ with $T_o$ frames, the first branch learns a feature $F(x)\in \mathbb{R} ^{H \times W \times T \times D}$, where each $D$-dimensional vector in $F(x)$ corresponds to a specific spatio-temporal region in the video.
The second branch generates $P$ regions of interest, $M_{p_k^c}(x) \in \mathbb{R} ^{H \times W \times T}$, that specify which regions of $F(x)$ are relevant for comparing with each prototype $p_k^c$.

The features from different spatio-temporal regions must be pooled before being compared to prototypes. As in~\cite{XprotoNet}, we perform a weighted average pooling with the learned regions of interest as follows:
\begin{equation}
\label{eqn:occurrence_pool}
\begin{aligned}
     f_{p_k^c}(x) =\frac{1}{HWT} \sum_{H, W, T} |M_{p_k^c}(x)| \circ F(x) ,
\end{aligned}
\end{equation}
where $|.|$ is the absolute value and $\circ$ is the Hadamard product.

\subsubsection{Prototype Pooling.}
The similarity score of a feature vector $f_{p_k^c}$ and prototype $p_k^c$ is calculated using cosine similarity, which is then shifted to $[0,1]$:
\begin{equation}
\label{eqn:cosine_sim}
\begin{aligned}
     g(x,p_k^c) =
     \frac{1}{2}
     (1 + 
     \frac{<f_{p_k^c}(x),p_k^c>}{\lVert f_{p_k^c}(x) \rVert_2 \lVert p_k^c \rVert_2} ) .
\end{aligned}
\end{equation}

\subsubsection{Prototypes for Aleatoric Uncertainty Estimation.}
In Fig.~\ref{fig:network}, trainable uncertainty prototypes (denoted $p_k^u$) are added to capture regions in the data distribution that are inherently ambiguous (Fig.~\ref{fig:network}.B). 
We use similarity between $f_{p_k^u}(x)$ and $p_k^u$ to quantify aleatoric uncertainty, denoted $\alpha \in [0,1]$.
We use an ``abstention loss'' (Eq.~(\ref{eqn:abstention_loss})) method inspired by~\cite{devries2018learning} to learn $\alpha$ and thereby $p_k^u$. 
In this loss, $\alpha$ is used to interpolate between the ground truth and prediction, pushing the model to ``abstain'' from its own answer at a penalty. 
\begin{align}
    \hat{y} & = \sigma (h (g(x, p_k^c))), \quad 
    \alpha  = \sigma (h (g(x, p_k^u))) ; \\
    \hat{y}' & = (1-\alpha) \hat{y} + \alpha y ; \\ 
    \mathcal{L}_{abs} & = CrsEnt(\hat{y}', y) - \lambda_{abs} \log (1 - \alpha) , 
    \label{eqn:abstention_loss}
\end{align}
where $\sigma$ denotes Softmax normalization in the output of $h(.)$, $y$ and $\hat{y}$ are the ground truth and the predicted probabilities, respectively, and $\lambda_{abs}$ is a regularization constant. 

When projecting $p_k^u$ to the nearest extracted feature from training examples, we relax the requirement in Eq.~(\ref{eqn:proto_push})  
allowing the uncertainty prototypes to be pushed to data with the ground truth of any AS severity class.

\subsubsection{Class-Wise Similarity Score.} 
The fully connected (FC) layer $h(.)$ is a dense mapping from prototype similarity scores to prediction logits. Its weights, $w_h$, are initialized to be 1 between class $c$ and the corresponding prototypes and 0 otherwise to enforce the process to resemble positive reasoning. 
$h(.)$ produces a score for membership in each class and for $\alpha$. 

\subsubsection{Loss Function.} 
As in previous prototype-based methods~\cite{chen2019ProtoPNet,XprotoNet}, the following losses are introduced to improve performance: 1) Clustering and separation losses (Eq.~(\ref{eqn:cluster_loss})), which encourage clustering based on class, where $\mathcal{P}_y$ denotes the set of prototypes belonging to class $y$.
Due to lack of ground truth uncertainties, these losses are only measured on $p_k^c$, not $p_k^u$; 2) Orthogonality loss (Eq.~(\ref{eqn:orth_loss})), which encourages prototypes to be more diverse; 3) Transformation loss $\mathcal{L}_{trns}$ (described in \cite{XprotoNet}), which regularizes the consistency of the predicted occurrence regions under random affine transformations; 
4) Finally, $\mathcal{L}_{norm}$ (described in \cite{chen2019ProtoPNet}) regularizes $w_h$ to be close to its initialization and penalizes relying on similarity to one class to influence the logits of other classes. 
Eq.~(\ref{eqn:overall_loss}) describes the overall loss function where $\lambda$ represent regularization coefficients for each loss term. The network is trained end-to-end. We conduct a ``push'' stage (see Eq.~(\ref{eqn:proto_push})) every 5 epochs to ensure that the learned prototypes are consistent with the embeddings from real examples.
\begin{align}
    \mathcal{L}_{clst} & = - \max_{p_k^c \in \mathcal{P}_y} g(x, p_k^c), \quad 
    \mathcal{L}_{sep} = \max_{p_k^c \notin \mathcal{P}_y} g(x, p_k^c); \label{eqn:cluster_loss} \\
    \mathcal{L}_{orth} & = \sum_{i > j} \frac{<p_i,p_j>}{\lVert p_i \rVert_2 \lVert p_j \rVert_2}; \label{eqn:orth_loss} \\
    \mathcal{L} = \mathcal{L}_{abs} & + 
    \lambda_{clst} \mathcal{L}_{clst} + 
    \lambda_{sep} \mathcal{L}_{sep} +
    \lambda_{orth} \mathcal{L}_{orth} +
    \lambda_{trns} \mathcal{L}_{trns} + 
    \lambda_{norm} \mathcal{L}_{norm} .
    \label{eqn:overall_loss}
\end{align}

\section{Experiments and Results}
\subsection{Datasets}
We conducted experiments on a private AS dataset and the public TMED-2 dataset~\cite{huangTMED2Dataset2022}. 
The private dataset was extracted from an echo study database of a tertiary care hospital with institutional review ethics board approval. 
Videos were acquired with Philips iE33, Vivid i, and Vivid E9 ultrasound machines.
For each study, the AS severity was classified using clinically standard Doppler echo guidelines~\cite{bonow2006acc} by a level III echocardiographer, keeping only cases with concordant Doppler measurements. 
PLAX and PSAX view cines were extracted from each study using a view-detection algorithm~\cite{liao2019modelling}, and subsequently screened by a level III echocardiographer to remove misclassified cines.
For each cine, the echo beam area was isolated and image annotations were removed.
The dataset consists of 5055 PLAX and 4062 PSAX view cines, with a total of 2572 studies. 
These studies were divided into training, validation, and test sets, ensuring patient exclusivity and following an 80-10-10 ratio.
We performed randomized augmentations including resized cropping and rotation.

The TMED-2 dataset~\cite{huangTMED2Dataset2022} consists of 599 fully labeled echo studies containing 17270 images in total. 
Each study consists of 2D echo images with clinician-annotated view labels (PLAX/PSAX/Other) and Doppler-derived study-level AS severity labels (no AS/early AS/significant AS).
Though the dataset includes an unlabeled portion, we trained on the labeled set only.
We performed data augmentation similar to the private dataset without time-domain operations.

\subsection{Implementation Details}
To better compare the results with TMED-2 dataset, we adopted their labeling scheme of no AS (normal), early AS (mild), and significant AS (moderate and severe) in our private dataset.
We split longer cines into 32-frame clips which are approximately one heart cycle long.
In both layers of the feature module, we used $D$ convolutional filters, while the three layers in the ROI module had $D$, $\frac{D}{2}$, and $P$ convolutional filters, preventing an abrupt reduction of channels to the relatively low value of $P$.
In both modules, we used kernel size of 1×1×1.
We set $D=256$ and $K=10$ for AS class and aleatoric uncertainty prototypes.  
Derived from the hyperparameter selection of ProtoPNet~\cite{chen2019ProtoPNet}, we assigned the values of $0.8$, $0.08$, and $10^{-4}$ to $\lambda_{clst}$, $\lambda_{sep}$, and $\lambda_{norm}$ respectively.
Through a search across five values of $0.1$, $0.3$, $0.5$, $0.9$, and $1.0$, we found the optimal $\lambda_{abs}$ to be $0.3$ based on the mean F1 score of the validation set.
Additionally, we found $\lambda_{orth}$ and $\lambda_{trns}$ to be  empirically better as $10^{-2}$ and $10^{-3}$ respectively.
We implemented our framework in PyTorch and trained the model end-to-end on one 16~GB NVIDIA Tesla V100 GPU.

\subsection{Evaluations on Private Dataset}
\subsubsection{Quantitative Assessment.}
In Table~\ref{tab:results_comparison}, we report
the performance of ProtoASNet in AS severity classification against the black-box baselines for image (Huang et al.~\cite{huangSemisupervisedEchocardiogramBenchmark2021}), video (Ginsberg et al.~\cite{AS_Tom}), as well as other prototypical methods, i.e. ProtoPNet~\cite{chen2019ProtoPNet} and XProtoNet~\cite{XprotoNet}.
In particular, for ProtoASNet, ProtoPNet~\cite{chen2019ProtoPNet}, and XProtoNet~\cite{XprotoNet}, we conduct both image-based and video-based experiments with ResNet-18 and R(2+1)D-18 backbones respectively.
We apply softmax to normalize the ProtoASNet output scores, including $\alpha$, to obtain class probabilities that account for the presence of aleatoric uncertainty.
We aggregate model predictions by averaging their probabilities from the image- (or clip-) level to obtain cine- and study-level predictions.
We believe the uncertainty probabilities reduce the effect of less informative datapoints on final aggregated results.
Additionally, the video-based models perform better than the image-based ones because the learnt prototypes can also capture AV motion which is an indicator of AS severity.
These two factors may explain why our proposed method, ProtoASNet, outperforms all other methods for study-level classification.

\subsubsection{Qualitative Assessment.}
\begin{figure}[t]
\centering
\includegraphics[width=0.99\textwidth]{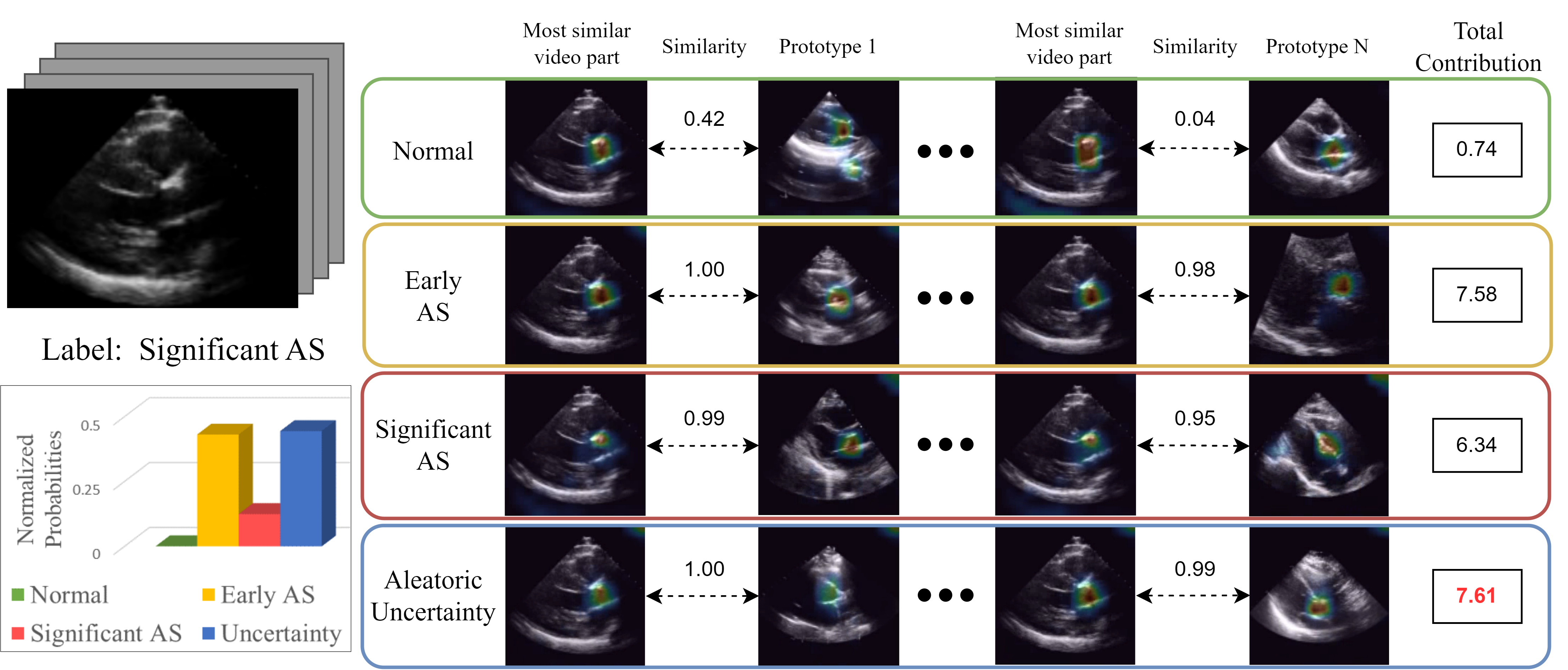}
\caption{
Visualization of the ProtoASNet decision-making process for a test cine video showing significant AS but poor valve leaflet visualization. 
We visualize most similar video parts by overlaying the upsampled model-generated ROI, $M_{p_k^c}(x_{test})$, on the test cine video.
Likewise, we visualize prototypes by finding the training clip each prototype is drawn from, $x_p$, and overlaying $M_{p_k^c}(x_p)$.
ProtoASNet explains which spatio-temporal parts of the test echo are most similar to the prototypes and how accumulation of these supporting evidence results in the prediction probabilities.
More visualizations of our model's performance are included in the supplementary material in video format. 
}

\label{fig:local_exp}
\end{figure}

The interpretable reasoning process of ProtoASNet for a video example is shown in Fig.~\ref{fig:local_exp}.
We observe that ProtoASNet places significant importance on prototypes corresponding to thickened AV leaflets due to calcification, which is a characteristic of both early and significant AS. 
Additionally, prototypes mostly capture the part of the heart cycle that aligns with the opening of the AV, providing a clinical indication of how well the valve opens up to be able to pump blood to the rest of the body.
This makes ProtoASNet's reasoning process interpretable for the user. 
Note how the uncertainty prototypes focusing on AV regions where the valve leaflets are not visible, are contributing to the uncertainty measure, resulting in the case being flagged as uncertain. 

\subsubsection{Ablation Study.}
We assessed the effect of removing distinct components of our design: uncertainty prototypes ($\mathcal{L}_{abs}, p_k^u$), clustering and separation ($\mathcal{L}_{clst}, \mathcal{L}_{sep}$), and \textit{push} mechanism. 
As shown in Table~\ref{tab:ablation}, keeping all the aforementioned components results in superior performance in terms of bACC and bMAE.
We evaluated whether the model is capable of detecting its own misclassification using the value of $\alpha$ (or entropy of the class predictions in the case without $\mathcal{L}_{abs}, p_k^u$).
This is measured by the AUROC of detecting ($y \neq \hat{y}$).
Learning $p_k^u$ may benefit accuracy by mitigating the overfitting of $p_k^c$ to poor-quality videos.
Furthermore, $\alpha$ seems to be a stronger indicator for misclassification than entropy.
Moreover, we measured prototype quality using diversity and sparsity~\cite{hesse2022insightr}, normalized by the total number of prototypes.
Ideally, each prediction can be explained by a low number of prototypes (low $s_{spars}$) but different predictions are explained with different prototypes (high Diversity).
When $\mathcal{L}_{clst}$ and $\mathcal{L}_{sep}$ are removed, the protoypes are less constrained, which contributes to stronger misclassification detection and more diversity, but reduce accuracy and cause explanations to be less sparse.
Finally, the \textit{push} mechanism improves performance, countering the intuition of an interpretability-performance trade-off.

\begin{table}[t]
\caption{Quantitative results on the test set of our private dataset in terms of balanced accuracy (bACC), mean F1 score, and balanced mean absolute error (bMAE). bMAE is the average of the MAE of each class, assuming labels of $0, 1, 2$ for no AS, early AS and significant AS respectively. Study-level results were calculated by averaging the prediction probabilities over all cines of each study. 
Results are shown as "mean(std)" calculated across five repetitions for each experiment.
Best results are in bold.
}
\label{tab:results_comparison}
\centering
\begin{tabular}{c|ccc|ccc}
\multirow{2}{*}{Method} & \multicolumn{3}{c}{Cine-level (N=973)}                                    
& \multicolumn{3}{c}{Study-level (N=258)}                      
\\ 
& \multicolumn{1}{c}{bACC$\uparrow$} 
& \multicolumn{1}{c}{F1 $\uparrow$} 
& \multicolumn{1}{c}{bMAE$\downarrow$}
& \multicolumn{1}{c}{bACC$\uparrow$} 
& \multicolumn{1}{c}{F1 $\uparrow$} 
& \multicolumn{1}{c}{bMAE$\downarrow$} \\ 
\hline 
\hline
Huang et al.~\cite{huangTMED2Dataset2022} 
& $70.2(1.5)$
& $0.70(.02)$
& $0.33(.02)$
& $74.7(1.6)$
& $0.75(.02)$
& $0.28(.02)$  \\
ProtoPNet~\cite{chen2019ProtoPNet} 
& $67.8(3.7)$
& $0.66(.05)$
& $0.36(.05)$
& $70.9(4.7)$
& $0.69(.07)$
& $0.32(.05)$   \\ 
XProtoNet~\cite{XprotoNet}
& $69.2(1.3)$
& $0.69(.01)$
& $0.34(.01)$
& $73.8(0.8)$  
& $0.74(.01)$   
& $0.29(.01)$ \\
ProtoASNet (Image)*
& $70.1(1.6)$
& $0.70(.02)$
& $0.33(.02)$
& $73.9(3.5)$
& $0.74(.04)$
& $0.29(.04)$ \\ \hline
Ginsberg et al.~\cite{AS_Tom}
& $\mathbf{76.0(1.4)}$ 
& $\mathbf{0.76(.01)}$  
& $\mathbf{0.26(.01)}$
& $78.3(1.6)$ 
& $0.78(.01)$
& $0.24(.02)$ \\
XProtoNet (Video)*   
& $74.1(1.1)$
& $0.74(.01)$
& $0.29(.01)$
& $77.2(1.4)$ 
& $0.77(.01)$  
& $0.25(.02)$ \\
ProtoASNet
& $75.4(0.9)$
& $0.75(.01)$
& $0.27(.01)$
& $\mathbf{ 80.0(1.1)}$
& $\mathbf{ 0.80(.01)}$  
& $\mathbf{ 0.22(.01)}$ \\
\multicolumn{7}{l}{* Feature extraction modified to the corresponding input type.}\\
\end{tabular}
\end{table}

\subsection{Evaluation on TMED-2, a Public Dataset}
We also applied our method to TMED-2, a public image-based dataset for AS diagnosis. 
Consistent with~\cite{huangTMED2Dataset2022}, images were fed to a WideResNet-based prototype model with two output branches.
The view classifier branch used average-pooling of patches followed by a fully connected layer.
However, the AS diagnosis branch used the prototype setup outlined in Methods. 
A diagram of the overall architecture is available in the supplementary material.
We trained the model end-to-end with images from all views. 
During inference, images with high entropy in the predicted view and high aleatoric uncertainty for AS classification were discarded. Then, probabilities for PLAX and PSAX were used for weighted averaging to determine the study-level prediction.
Addition of the prototypical layer and thresholding on predicted uncertainty achieves 79.7\% accuracy for AS severity, outperforming existing black-box method~\cite{huangTMED2Dataset2022} at 74.6\%.

\section{Conclusion}
We introduce ProtoASNet, an interpretable method for classifying AS severity using B-mode echo that outperforms existing black-box methods. 
ProtoASNet identifies clinically relevant spatio-temporal prototypes that can be visualized to improve algorithmic transparency.
In addition, we introduce prototypes for estimating aleatoric uncertainty, which help flag difficult-to-diagnose scenarios, such as videos with poor visual quality. 
Future work will investigate methods to optimize the number of prototypes, or explore out-of-distribution detection using prototype-based methods.

\begin{table}[t]
\caption{Ablation study on the validation set of our private dataset.}
\label{tab:ablation}
\centering
\begin{tabular}{lc|ccccc}
\multicolumn{2}{c|}{\multirow{2}{*}{Method}} & \multicolumn{5}{c}{Clip-level (N=1280)}       \\
\multicolumn{2}{l|}{}                        & bACC $\uparrow$ & bMAE $\downarrow$  & $\text{AUROC}_{y \neq \hat{y}}\uparrow$ & $s_{spars} \downarrow$ & Diversity $\uparrow$ \\ \hline \hline
 & w/o $\mathcal{L}_{abs}, p_k^u$            
& 76.1 & 0.25 & 0.73       & 0.37     & \textbf{0.50}         \\
 & w/o $\mathcal{L}_{clst},\mathcal{L}_{sep}$             
& 74.8 & 0.26 & \textbf{0.79}       & 0.49     & \textbf{0.50}         \\
 & w/o \textit{push}             
& 77.9 & \textbf{0.23} & 0.75       & 0.35     & 0.43         \\
 & All parts (ours)              
& \textbf{78.4} & \textbf{0.23} & 0.75       & \textbf{0.33}    & 0.45        
\end{tabular}
\end{table}

\subsubsection{Acknowledgements.}
This work was supported in part by the Canadian Institutes of Health Research (CIHR) and in part by the Natural Sciences and Engineering Research Council of Canada (NSERC).

\bibliographystyle{splncs04}
\bibliography{ProtoASNet}

\begin{thebibliography}{10}
\providecommand{\url}[1]{\texttt{#1}}
\providecommand{\urlprefix}{URL }
\providecommand{\doi}[1]{https://doi.org/#1}

\bibitem{adebayo2018sanity}
Adebayo, J., Gilmer, J., Muelly, M., Goodfellow, I., Hardt, M., Kim, B.: Sanity
  checks for saliency maps. In: Advances in Neural Information Processing
  Systems. vol.~31. Curran Associates, Inc. (2018)

\bibitem{ancona2020epidemiology}
Ancona, R., Pinto, S.C.: Epidemiology of aortic valve stenosis (as) and of
  aortic valve incompetence (ai): Is the prevalence of as/ai similar in
  different parts of the world. European Society of Cardiology  \textbf{18}(10)
  (2020)

\bibitem{bonow2006acc}
Bonow, R.O., Carabello, B.A., Chatterjee, K., De~Leon, A.C., Faxon, D.P.,
  Freed, M.D., Gaasch, W.H., Lytle, B.W., Nishimura, R.A., O’Gara, P.T.,
  et~al.: Acc/aha 2006 guidelines for the management of patients with valvular
  heart disease: a report of the american college of cardiology/american heart
  association task force on practice guidelines (writing committee to revise
  the 1998 guidelines for the management of patients with valvular heart
  disease) developed in collaboration with the society of cardiovascular
  anesthesiologists endorsed by the society for cardiovascular angiography and
  interventions and the society of thoracic surgeons. Journal of the American
  College of Cardiology  \textbf{48}(3),  e1--e148 (2006)

\bibitem{chen2019ProtoPNet}
Chen, C., Li, O., Tao, D., Barnett, A., Rudin, C., Su, J.K.: This looks like
  that: deep learning for interpretable image recognition. Advances in neural
  information processing systems  \textbf{32} (2019)

\bibitem{devries2018learning}
DeVries, T., Taylor, G.W.: Learning confidence for out-of-distribution
  detection in neural networks. arXiv preprint arXiv:1802.04865  (2018)

\bibitem{AS_Tom}
Ginsberg, T., Tal, R.e., Tsang, M., Macdonald, C., Dezaki, F.T., van~der Kuur,
  J., Luong, C., Abolmaesumi, P., Tsang, T.: Deep video networks for automatic
  assessment of aortic stenosis in echocardiography. In: Noble, J.A., Aylward,
  S., Grimwood, A., Min, Z., Lee, S.L., Hu, Y. (eds.) Simplifying Medical
  Ultrasound: Second International Workshop, Strasbourg, France, September 27,
  2021, Proceedings 2. pp. 202--210. Springer (2021)

\bibitem{gulshad2023hierarchical}
Gulshad, S., Long, T., van Noord, N.: Hierarchical explanations for video
  action recognition. arXiv e-prints pp. arXiv--2301 (2023)

\bibitem{hesse2022insightr}
Hesse, L.S., Namburete, A.I.: Insightr-net: Interpretable neural network for
  regression using similarity-based comparisons to prototypical examples. In:
  Medical Image Computing and Computer Assisted Intervention--MICCAI 2022: 25th
  International Conference, Singapore, September 18--22, 2022, Proceedings,
  Part III. pp. 502--511. Springer (2022)

\bibitem{huangSemisupervisedEchocardiogramBenchmark2021}
Huang, Z., Long, G., Wessler, B., Hughes, M.C.: A new semi-supervised learning
  benchmark for classifying view and diagnosing aortic stenosis from
  echocardiograms. In: Proceedings of the 6th Machine Learning for Healthcare
  Conference (2021)

\bibitem{huangTMED2Dataset2022}
Huang, Z., Long, G., Wessler, B., Hughes, M.C.: Tmed 2: A dataset for
  semi-supervised classification of echocardiograms. In: DataPerf: Benchmarking
  Data for Data-Centric AI Workshop (2022)

\bibitem{XAI_RegionGrouping}
Huang, Z., Li, Y.: Interpretable and accurate fine-grained recognition via
  region grouping. In: Proceedings of the IEEE/CVF Conference on Computer
  Vision and Pattern Recognition. pp. 8662--8672 (2020)

\bibitem{XprotoNet}
Kim, E., Kim, S., Seo, M., Yoon, S.: Xprotonet: Diagnosis in chest radiography
  with global and local explanations. Proceedings of the IEEE/CVF Conference on
  Computer Vision and Pattern Recognition pp. 15714--15723 (2021)

\bibitem{liao2019modelling}
Liao, Z., Girgis, H., Abdi, A., Vaseli, H., Hetherington, J., Rohling, R., Gin,
  K., Tsang, T., Abolmaesumi, P.: On modelling label uncertainty in deep neural
  networks: automatic estimation of intra-observer variability in 2d
  echocardiography quality assessment. IEEE Transactions on Medical Imaging
  \textbf{39}(6),  1868--1883 (2019)

\bibitem{minners2008inconsistencies}
Minners, J., Allgeier, M., Gohlke-Baerwolf, C., Kienzle, R.P., Neumann, F.J.,
  Jander, N.: Inconsistencies of echocardiographic criteria for the grading of
  aortic valve stenosis. European Heart Journal  \textbf{29}(8),  1043--1048
  (2008)

\bibitem{minners2010inconsistent}
Minners, J., Allgeier, M., Gohlke-Baerwolf, C., Kienzle, R.P., Neumann, F.J.,
  Jander, N.: Inconsistent grading of aortic valve stenosis by current
  guidelines: haemodynamic studies in patients with apparently normal left
  ventricular function. Heart  \textbf{96}(18),  1463--1468 (2010)

\bibitem{acc_guidelines20212020}
Otto, C.M., Nishimura, R.A., Bonow, R.O., Carabello, B.A., Erwin~III, J.P.,
  Gentile, F., Jneid, H., Krieger, E.V., Mack, M., et~al.: 2020 acc/aha
  guideline for the management of patients with valvular heart disease: a
  report of the american college of cardiology/american heart association joint
  committee on clinical practice guidelines. American College of Cardiology
  Foundation Washington DC  \textbf{77}(4),  e25--e197 (2021)

\bibitem{rudin2019stop}
Rudin, C.: Stop explaining black box machine learning models for high stakes
  decisions and use interpretable models instead. Nature Machine Intelligence
  \textbf{1}(5),  206--215 (2019)

\bibitem{GradCam}
Selvaraju, R.R., Cogswell, M., Das, A., et~al.: Grad-cam: Visual explanations
  from deep networks via gradient-based localization. In: Proceedings of the
  IEEE International Conference on Computer Vision (Oct 2017)

\bibitem{journals/corr/SimonyanVZ13}
Simonyan, K., Vedaldi, A., Zisserman, A.: Deep inside convolutional networks:
  Visualising image classification models and saliency maps. 2nd International
  Conference on Learning Representations, Workshop Track Proceedings  (2014)

\bibitem{thoenes2018patient}
Thoenes, M., Bramlage, P., Zamorano, P., Messika-Zeitoun, D., Wendt, D., Kasel,
  M., Kurucova, J., Steeds, R.P.: Patient screening for early detection of
  aortic stenosis (as)—review of current practice and future perspectives.
  Journal of Thoracic Disease  \textbf{10}(9), ~5584 (2018)

\bibitem{tran2018closer}
Tran, D., Wang, H., Torresani, L., Ray, J., LeCun, Y., Paluri, M.: A closer
  look at spatiotemporal convolutions for action recognition. In: Proceedings
  of the IEEE conference on Computer Vision and Pattern Recognition. pp.
  6450--6459 (2018)

\bibitem{Trinh_2021_WACV}
Trinh, L., Tsang, M., Rambhatla, S., Liu, Y.: Interpretable and trustworthy
  deepfake detection via dynamic prototypes. Proceedings of the IEEE/CVF Winter
  Conference on Applications of Computer Vision pp. 1973--1983 (2021)

\bibitem{wang2021TesNet}
Wang, J., Liu, H., Wang, X., Jing, L.: Interpretable image recognition by
  constructing transparent embedding space. Proceedings of the IEEE/CVF
  International Conference on Computer Vision pp. 895--904 (2021)

\end{thebibliography}

\end{document}


\title{Supplementary Material}


\author{Hooman Vaseli et al.}

\authorrunning{Hooman Vaseli et al.}

\institute{Department of Electrical and Computer Engineering, The University of British Columbia, Vancouver, BC, Canada}

\maketitle              

\begin{figure}[h]
\centering
\includegraphics[width=0.99\textwidth]{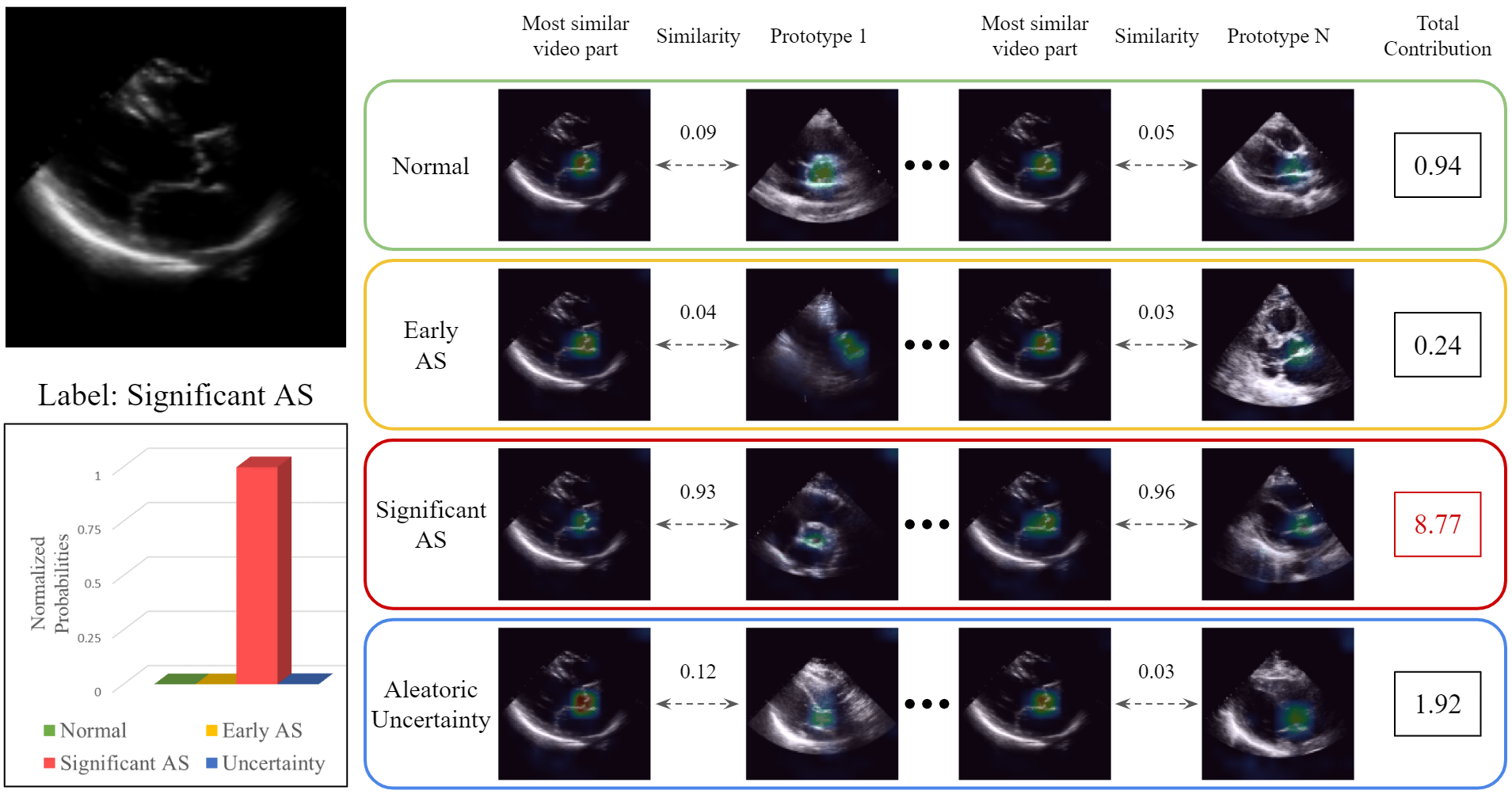}
\caption{
Visualization of the ProtoASNet decision-making process for a test cine video showing significant AS. 
Mp4 video clip of this visualization is included in the supplementary material.
}
\end{figure}

\begin{figure}[h!]
    \centering
    \includegraphics[width=\linewidth]{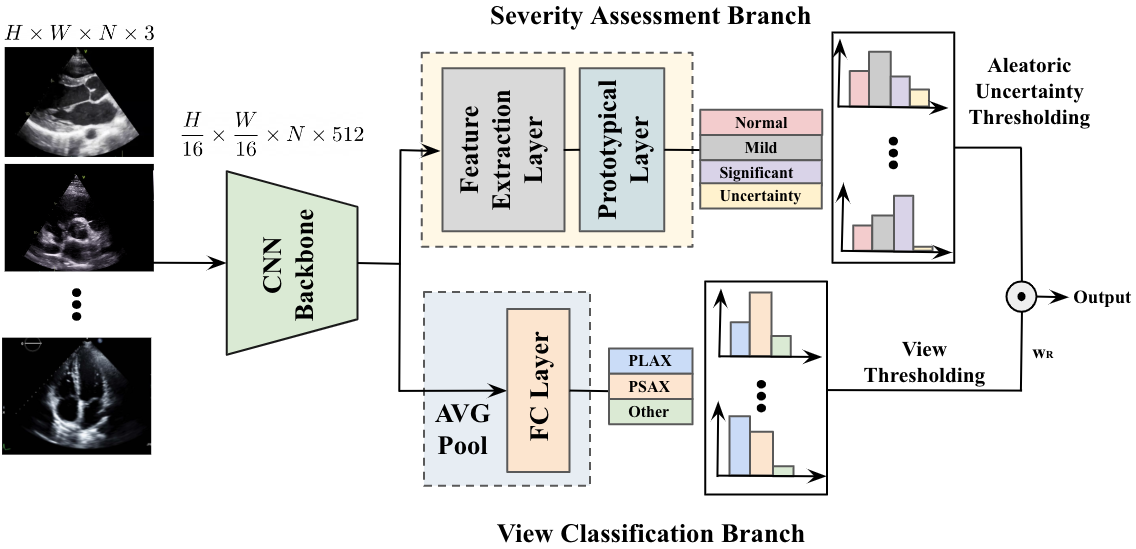}
    \caption{\textbf{Study-level diagnosis of TMED2 dataset} - Given image-level trained network we get the study-level diagnosis by aggregating images with low view classification entropy and uncertainty probability while prioritizing diagnoses from images with PLAX and PSAX view.}
    \label{fig:attn_sup}
\end{figure}

\begin{table}[h]
\caption{Diagnostic performance on the test set of the TMED-2 dataset.
Image predictions are derived with  probabilities with prioritized view averaging and thresholding.}
\label{tab:tmed_results}
\resizebox{\textwidth}{!}{
\begin{tabular}{c|cccc|cc}
\multirow{2}{*}{Method} & \multicolumn{4}{c|}{Aggregation Method}
& \multicolumn{2}{c}{Study-level}                      
\\ 
& \multicolumn{1}{c}{Proto Layer} 
& \multicolumn{1}{c}{Prioritized View} 
& \multicolumn{1}{c}{View Thresh} 
& \multicolumn{1}{c|}{Abs Thresh} 
& \multicolumn{1}{c}{View BACC} 
& \multicolumn{1}{c}{AS BACC}  \\ 
\hline 
\hline
Huang et al
& x
& \checkmark 
& \checkmark 
& x 
& 96.2\%
& 74.6\% \\ 
Ours 
& \checkmark 
& \checkmark 
& \checkmark 
& \checkmark 
& 96.1\%  
& \textbf{79.7\%}   \\ 
\end{tabular}
}
\end{table}

\begin{table}[h!]
    \settowidth\rotheadsize{Example 1}
    \begin{tabular}
    {@{\hspace{0mm}}c@{\hspace{1mm}}c@{\hspace{1mm}}c@{\hspace{1mm}}c}
    \centering
        \rothead{\centering Healthy} &
        \includegraphics[width=0.29\textwidth,trim={0 0 0 0},clip,valign=m]{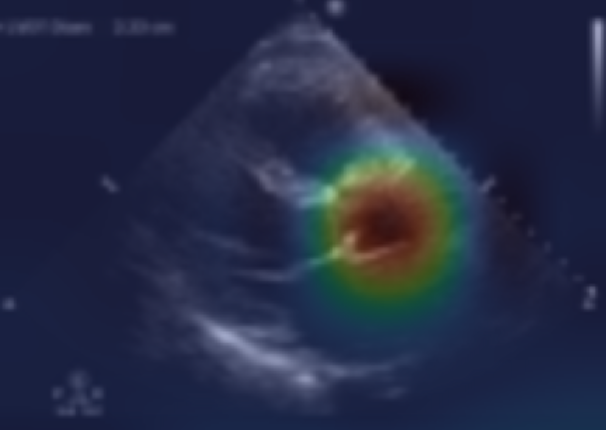} & 
        \includegraphics[width=0.29\textwidth,trim={0 0 0 0},clip,valign=m]{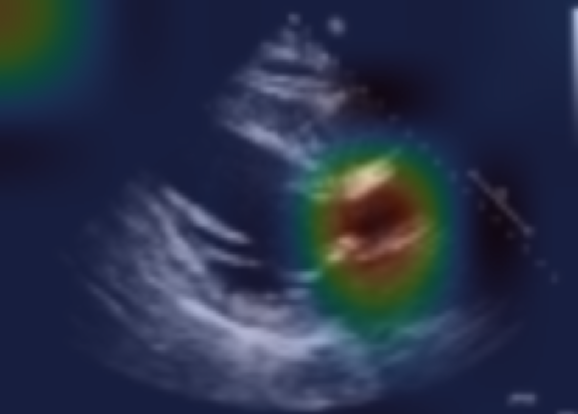} & 
        \includegraphics[width=0.29\textwidth,trim={0 0 0 0},clip,valign=m]{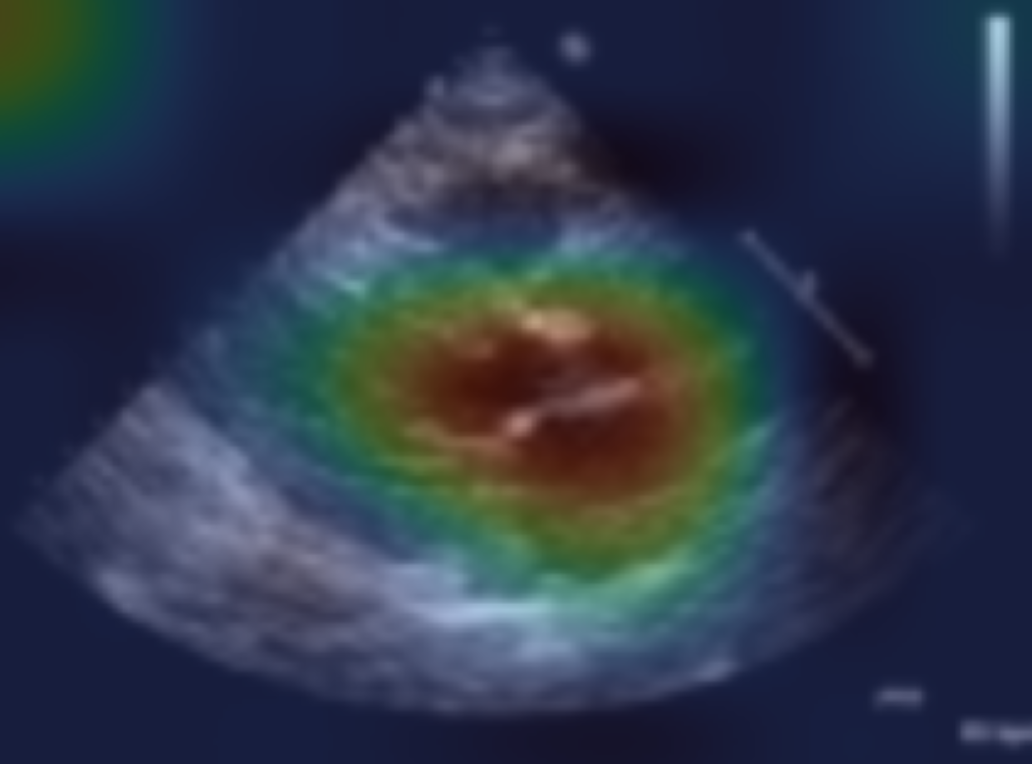} 
        \\ \addlinespace[0.5mm]
        \rothead{\centering Significant} &
        \includegraphics[width=0.29\textwidth,trim={0 0 0 0},clip,valign=m]{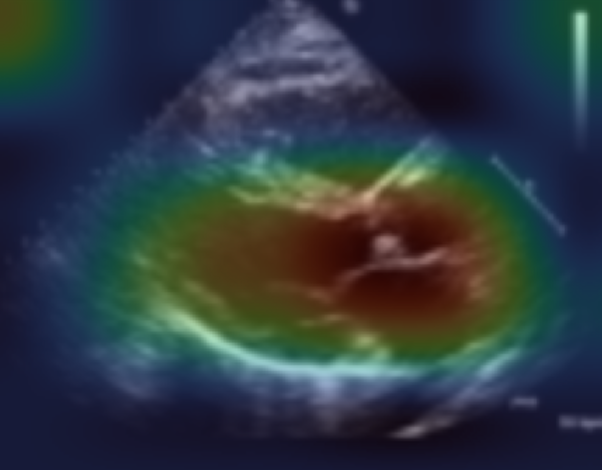} & 
        \includegraphics[width=0.29\textwidth,trim={0 0 0 0},clip,valign=m]{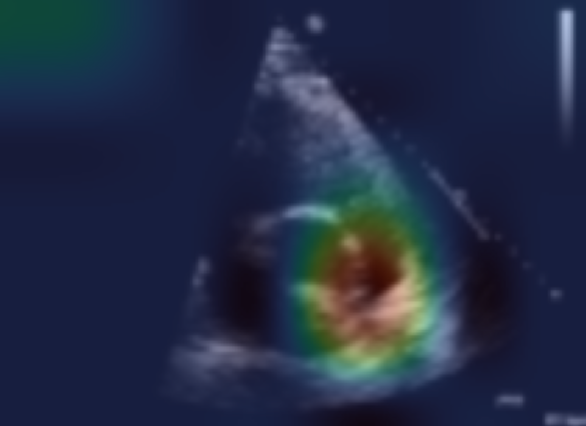} & 
        \includegraphics[width=0.29\textwidth,trim={0 0 0 0},clip,valign=m]{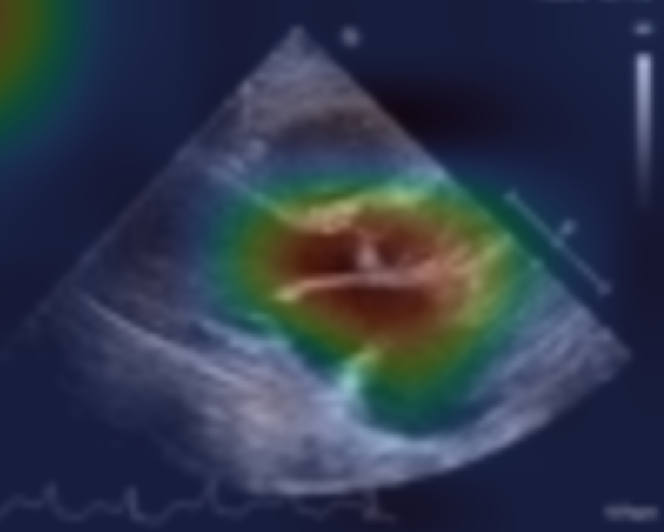} 
        \\ \addlinespace[0.5mm]
        \rothead{\centering Uncertainty} &
        \includegraphics[width=0.29\textwidth,trim={0 0 0 0},clip,valign=m]{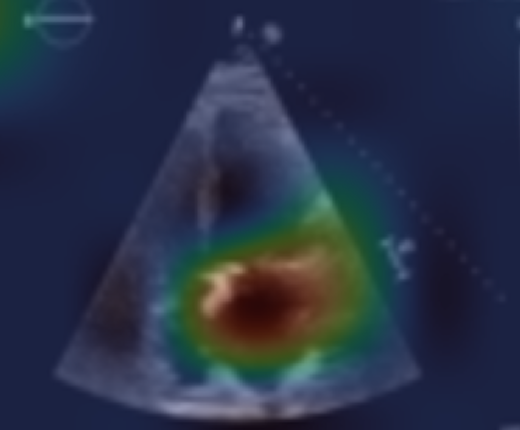} & 
        \includegraphics[width=0.29\textwidth,trim={0 0 0 0},clip,valign=m]{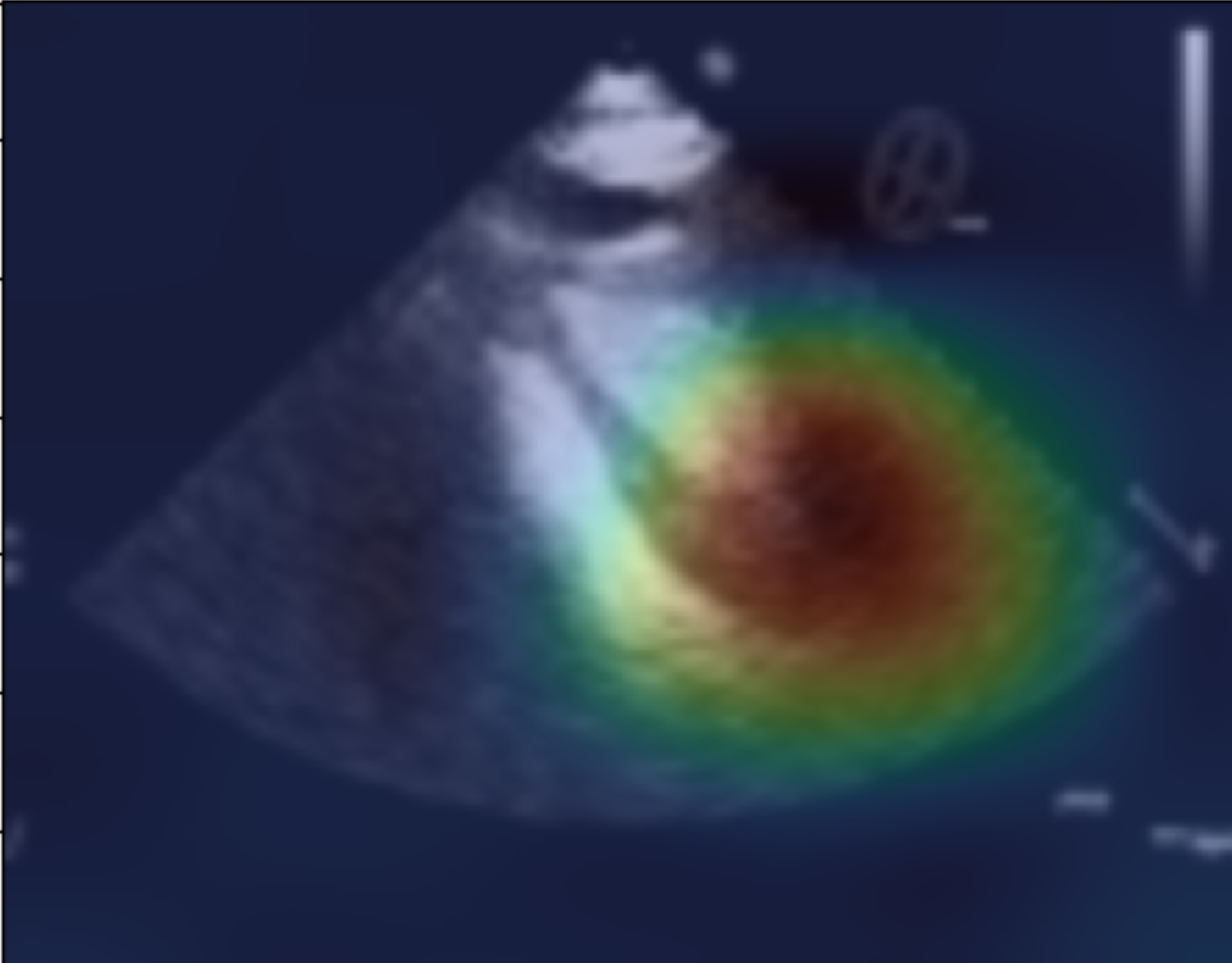} & 
        \includegraphics[width=0.29\textwidth,trim={0 0 0 0},clip,valign=m]{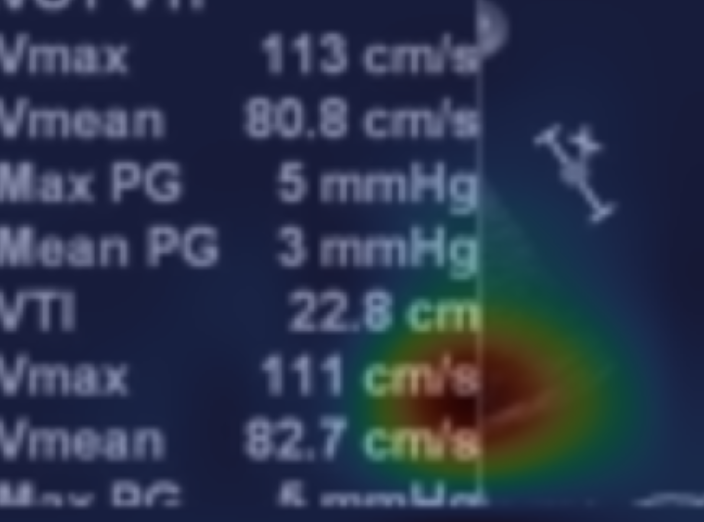} 
        \\ \addlinespace[0.5mm]
    \end{tabular}
    \caption{ Most prototypes focus on areas around the aortic valve. Three example prototypes for healthy and significant AS have been demonstrated. Three uncertainty prototypes are also shown. Most uncertainty prototypes capture views other than PLAX and PSAX or images that contain extra information such as EKG or patient information.}
     \label{tab:tmed_representations}
\end{table}